\documentclass[10pt,letterpaper,twocolumn]{article}
\usepackage[numbers]{preprint}
\usepackage{xspace,pifont,adjustbox,placeins}
\usepackage{xcolor}
\usepackage[normalem]{ulem}
\newcommand{\methodname}{OpenTac\xspace}
\newcommand{\coolname}{\methodname}
\newcommand{\cadname}{OpenTac-CAD\xspace}
\newcommand{\simname}{OpenTac-Sim\xspace}
\newcommand{\transfername}{OpenTac-Transfer\xspace}
\newcommand{\xamplingname}{OpenTac-Xampling\xspace}
\newcommand{\etal}{\emph{et al.}}
\title{OpenTac: An Open Ecosystem for Customizable, Scalable, and Transferable 3D Tactile Sensing}
\author{%
Ariel Slepyan$^{1,2}$, Geeta Chandra Raju Bethala$^1$, Jiafeng Gu$^1$,
Anagha Rajendra Dangle$^1$, Yinzhe Zhou$^1$, Hongyuan Fu$^1$,
Jade Sim$^1$, Tuhin Basu$^1$, Xinyi Yin$^1$,
Md Faizal Karim$^1$,
Gregory Hager\textsuperscript{$\dagger 1$},
Homanga Bharadhwaj\textsuperscript{$\dagger 1$},
Anand Bhattad\textsuperscript{$\dagger 1$},
Krishna Murthy Jatavallabhula\textsuperscript{$\dagger 1$},
Nitish V. Thakor\textsuperscript{$\dagger 1$}%
}

\affiliations{%
\large{$^1$ Johns Hopkins University, $^2$ Massachusetts Institute of Technology}\\[3pt]
{\small\textsuperscript{$\dagger$}Equal advising}\\
\Large{ \href{https://opentac.github.io/}
        {\texttt{\textbf{opentac.github.io}}}}
}
\runningtitle{OpenTac}
\preprintnotice{Preprint}
\hypersetup{pdftitle={OpenTac: An Open Ecosystem for Customizable, Scalable, and Transferable 3D Tactile Sensing},pdfauthor={}}

\teaser{    \includegraphics[width=\linewidth]{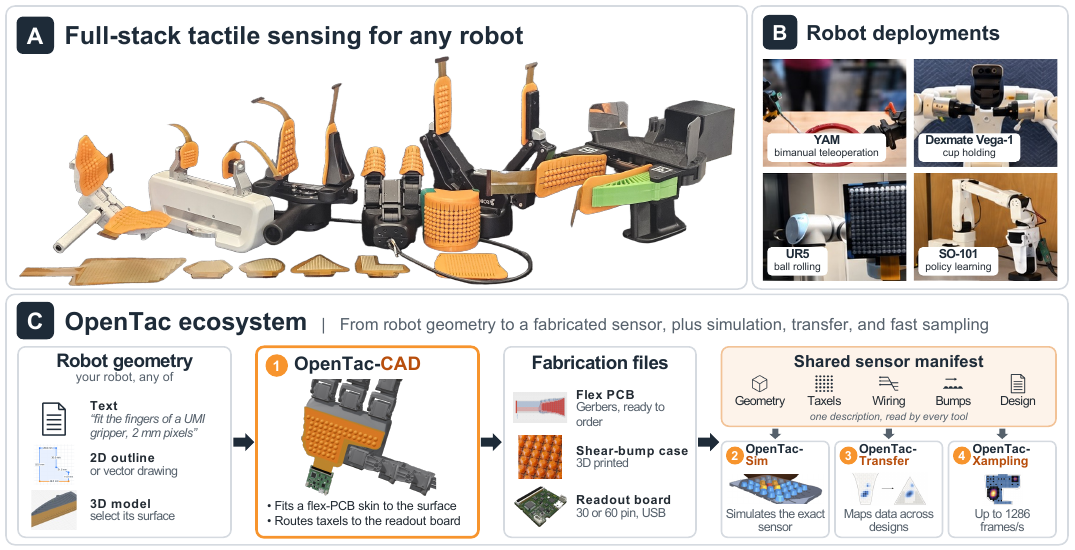}}{
\textbf{The \coolname ecosystem.}
\textbf{(A)} Custom tactile skins that sense both normal and shear force, built for robot hands, grippers, arms, and feet. \textbf{(B)} Robot deployments on 4 real-world platforms.
\textbf{(C)} \ding{182}~\cadname compiles a text description, 2D drawing, or 3D CAD surface into a fabrication-ready flex PCB, a shear-bump case, and a machine-readable sensor manifest.
The manifest is shared by \ding{183}~\simname, which instantiates the corresponding sensor geometry in simulation; \ding{184}~\transfername, which maps tactile observations between sensor designs; and \ding{185}~\xamplingname, which performs adaptive partial sampling at up to kilohertz rates.
}{fig:overview}

\wideabstract{
We present \coolname, a novel sensor and a full-stack ecosystem for tactile skins that are highly customizable, scalable, and readily deployed in downstream robot control and learning applications. Our core contribution is twofold. First, we develop a scalable resistive tactile skin with a structured deformable overlay that converts tangential loading into spatially varying pressure.
Differential responses across neighboring `taxels' support decoding of local shear vectors, and the common-mode response encodes normal force. Second, we provide a highly customizable sensor design tool that is able to automatically generate fabrication files for these tactile skins from 2D outlines and developable 3D surfaces, including from abstract descriptions such as text. 
We develop an ecosystem around the OpenTac sensor, including the CAD design tool, a sensor simulator, cross-sensor transfer, and fast/adaptive data acquisition.
We demonstrate the broad applicability of the OpenTac-CAD sensor design tool by designing and fabricating more than 30 sensors across a variety of robot embodiments.
We characterize the three-dimensional tactile fields produced by OpenTac with a commercial, high-precision force/torque sensor.
We demonstrate the broad applicability of 3D tactile sensing across contact-rich feedback control and manipulation policy learning on four real-world robot embodiments.
Additionally, adaptive acquisition enables kilohertz-rate partial readouts, and our readout circuitry enables high dynamic range (HDR) acquisition, measuring, for example, a 3.7~kg weight stack and a 20~g weight on the same sensor in the same frame.
We provide all hardware, software, and result videos at
\href{https://opentac.github.io/}
        {\texttt{\textbf{opentac.github.io}}}}

\begin{document}
\maketitle
\clearpage

\section{Introduction}
\label{sec:introduction}
 As robots expand to perform tasks in more general, varied, and dynamic environments, their ability to manipulate quickly, reliably, and safely will increasingly depend on directly sensing and reacting to contact---that is, on tactile sensing. 
\begin{figure*}[t]
    \centering
    \includegraphics[width=\linewidth]{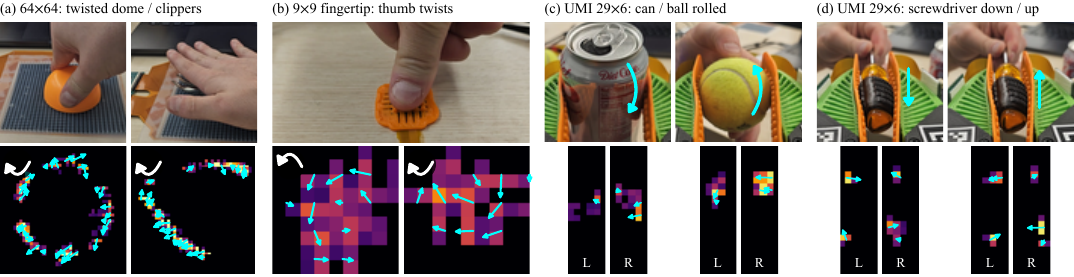}
    \caption{\textbf{Tactile samples recorded with OpenTac skins.} Heatmaps indicate normal load; cyan arrows denote per-bump shear; white arcs denote rotation estimates. (a) 64×64 skin, twisting a dome and clippers. (b) 9×9 fingertip, opposite thumb twists. (c) UMI gripper, 29×6 finger skins: can and ball rocked in the grasp. (d) Extrinsic contact: screwdriver moved down and up.}
    \label{fig:example-data}
\end{figure*}
However, sensors must first be customized to fit the end-effector or surface of interest. As a result, research on the use of tactile sensing is fundamentally limited by customizability: researchers must either adapt their work to sensors manufactured in fixed geometries, or they must possess sufficient design and fabrication expertise to develop custom sensors in-house. These constraints are particularly limiting when instrumenting custom-designed robot geometries or larger, irregular surfaces.

In this paper, we focus on \emph{conformal tactile skins}, as they integrate distributed sensing elements into thin, flexible surfaces that can be wrapped around robot geometries, enabling fully sensorized robot bodies. However, existing tactile skins, such as FlexiTac~\cite{flexitac}, provide only normal-force (1D) estimates and require manual redesign for each new geometry. We present a sensor design that can be easily customized to fit many robot geometries, scaled in sensing area or resolution, and used to estimate normal and shear forces (3D) over local groups of sensing elements.

Several efforts have partially addressed sensor customizability and made tactile skins more capable and accessible for the robot control and learning communities ~\cite{flexitac,anyskin,eflesh}. 
AnySkin provides replaceable sensing surfaces but relies on a predefined electronic layout~\cite{anyskin}, while eFlesh automates the design of a deformable sensor body but relies on separately supplied magnetometer electronics~\cite{eflesh}. %
In addition to the challenges of customizing the entire sensor, existing sensors are sampled at rates ranging between 30 and 100 Hz.
Such raster scanning-based acquisition can limit the ability to resolve and respond to contacts lasting only a few milliseconds. Beyond hardware, changes in sensor geometry must also be accommodated by the software used to acquire, simulate, and process tactile observations.

To address these limitations, we introduce \textbf{OpenTac}, an open sensor design environment and software ecosystem for customizable, scalable tactile sensing (Fig.~\ref{fig:overview}). OpenTac supports the design of low-cost piezoresistive tactile skins that estimate {\em both} local normal and shear forces with millimeter-scale spatial resolution over planar and developable surfaces. Our sensor combines a flexible printed circuit board and a piezoresistive film with a structured elastomeric overlay that redistributes lateral pressure across neighboring sensing elements, allowing local shear to be decoded while retaining normal-force information. The sensing layout, electrode geometry, and mechanical overlay can be customized to the surface geometry and force range of interest.

\looseness=-1 To support this degree of customizability, OpenTac includes a full-stack ecosystem of design, simulation, and interoperability tools. \textbf{OpenTac-CAD} is a design tool that automatically generates sensor fabrication files from a desired geometry. Users can provide a 2D outline, select a developable surface on a 3D robot model, or specify an abstract verbal description, such as ``\emph{design a star-shaped sensor that fits inside a circle with a diameter 5~cm}.'' 
In addition, OpenTac-CAD produces a machine-readable \emph{manifest} file describing the sensor's geometry, electrical addressing, and mechanical configuration. 
\textbf{OpenTac-Sim} provides tools to simulate the sensors and sensor-object interactions for robot manipulation tasks across any sensor design described by the manifest file.
\textbf{OpenTac-Transfer} enables cross-sensor transfer by mapping calibrated tactile observations across sensor layouts using their physical coordinates and local surface frames. These components allow sensors with different shapes, resolutions, and electrical layouts to share acquisition and processing tools and provide a common interface for downstream control and learning.

\looseness=-1
Scaling tactile coverage also increases the number of measurements required to read the sensor. To address this, we also provide \textbf{OpenTac-Xampling}, an adaptive acquisition method
which operates across different sensor outlines using adaptive partial scans. For example, on a $32 \times 32$ skin, adaptive acquisition of sparse contact events reaches 1286 Hz, more than a 19$\times$ increase over full-array scans (67 Hz) achieved by conventional acquisition~\cite{sundaram2019stag}.

\looseness=-1
 We evaluate all components of the \coolname ecosystem. We use \cadname to generate fabrication files for more than 30 sensor layouts (including robot fingers, grippers, palms, and feet), fabricate the corresponding sensors, and deploy them on four real robot embodiments.
 We further evaluate the 3D force estimates against a commercial force/torque sensor.
 Finally, we demonstrate that our sensor's normal and shear force estimates enable contact-rich feedback control and manipulation policy learning.

To summarize, our main contributions are:
\begin{itemize}
    \item \textbf{OpenTac: a full-stack ecosystem} for automated sensor design, fabrication, simulation, cross-sensor observation mapping, and kilohertz-rate adaptive acquisition.
    \item A customizable tactile skin that estimates \textbf{normal and shear forces} using a simple, scalable construction.
    \item \looseness=-1 Evaluations spanning over 30 fabricated layouts, sensor characterization, high-speed sensing, and real-world robot control and learning, demonstrating the utility of the resulting tactile observations for contact-rich manipulation.
\end{itemize}

All \coolname hardware, software, and firmware is available under permissive licenses, enabling replication, modification, and redistribution for research and commercial use.

\section{Related Work}
\label{sec:related_work}

\begin{table*}[t]
\centering
\begingroup
\newcommand{\otyes}{\ding{51}}
\newcommand{\otno}{\ding{55}}
\newcommand{\otmark}[1]{\textsuperscript{#1}}
\caption{Representative tactile sensing systems. Auto-design lists
automatically generated components. Shear denotes sensitivity to tangential force. Est. cost is per sensor unit (or sq. cm for skin). $^\dagger$ indicates cost per sq. cm of sensor, plus cost of a readout board.}
\label{tab:tactile_comparison}
\small
\setlength{\tabcolsep}{7pt}
\renewcommand{\arraystretch}{1.13}
\begin{adjustbox}{max width=\linewidth}
\begin{tabular}{@{}lcllrr@{}}
\toprule
\textbf{Sensor} & \textbf{Shear} & \textbf{Form} &
\textbf{Auto-design} & \shortstack{\textbf{Rate}\\\textbf{(Hz)}} &
\shortstack{\textbf{Est. cost}\\\textbf{(\$/unit)}} \\
\midrule
GelSight~\cite{gelsight} & \otyes & Finger & \otno & 30 & $65-400$ \\
DIGIT~\cite{digit360,lambeta2020digit} & \otyes (with markers) & Finger & \otno & 60 & $55-103$ \\
GelSlim 4.0~\cite{gelslim4} & \otyes & Finger & \otno & 30-60 & $73-123$ \\
TacTip~\cite{tactip} & \otyes & Finger & \otno & 30--120 & $67-169$ \\
uSkin~\cite{uskin} & \otyes & Patch & \otno & 100 & $10-25$ \\
ReSkin~\cite{reskin} & \otyes & Skin & \otno & 400 & $5.5-10.5$ \\
AnySkin~\cite{anyskin} & \otyes & Skin & Fingertip + Moulds & 100 & $2-10$ \\
eFlesh~\cite{eflesh} & \otyes & 3D body & Sensor surface & 100 & $25-50$ \\
iCub skin~\cite{icubskin} & \otno & Skin & \otno & 25 & $2.5-7$ \\
FlexiTac~\cite{flexitac} & \otno & Skin & \otno & 100 & $\approx 1.95 + 30^\dagger$ \\
Sun et al.~\cite{sun2019shear} & \otyes & Skin & \otno & N/A & $2-10$ \\
GenTact-Prox~\cite{kohlbrenner2026gentactprox} & \otno & Skin & \otyes & 15-30 & 25 \\
\midrule
\textbf{OpenTac} & \otyes & \textbf{Skin} &
\textbf{full-sensor} &
\shortstack{\textbf{67--1286}\\(adaptive)} & $\approx 1.95 + 80^\dagger$ \\
\bottomrule
\end{tabular}

\end{adjustbox}
\endgroup

\end{table*}

\looseness=-1
\textbf{Tactile sensing hardware} typically involves optical, magnetic, capacitive, and/or resistive transduction to recover contact. GelSight provides high-resolution optical contact sensing~\cite{gelsight}, while uSkin integrates
distributed three-axis force elements~\cite{uskin}. GelSlim uses marker displacements and inverse finite-element modeling to estimate dense contact forces~\cite{ma2019denseforce}, while Yan \etal~use a magnetized film to decouple normal force and one shear component~\cite{yan2021magneticskin}. Mechanical structures can also encode shear into differential pressure: Sun \etal~place a bump over four piezoresistive cells and infer
load direction from their relative responses~\cite{sun2019shear}. \methodname builds on this principle using a single-FPCB (flexible PCB) sensing architecture designed for automated customization and shared readout design that specifies a unified cross-sensor interface.

Early work on \textbf{customizing and scaling tactile skins} used modular or severable elements to cover complex
robot surfaces~\cite{ohmura2006,mittendorfer2011humanoid}, while later systems such as
the iCub skin, RobotSweater, and conformal tactile textiles target large-area conformable
coverage~\cite{icubskin,robotsweater,luo2021conformal}. ReSkin and AnySkin decouple the replaceable magnetic interface from the electronics~\cite{reskin,anyskin}. eFlesh automates magnetic microstructures~\cite{eflesh}, and FlexiTac provides an open flexible pressure array~\cite{flexitac}. Computational design tools have also produced procedurally generated 3D-printed whole-body skins~\cite{kohlbrenner2025gentact,kohlbrenner2026gentactprox}, personalized FPCB tactile gloves~\cite{murphy2025flexglove}, and surface-conforming capacitive sensors for existing 3D objects~\cite{ilic2026retrofit}. Lim \etal~use laser-patterned flexible circuits to integrate capacitive pressure and bending sensors with temperature and proximity modules in customizable multimodal skins~\cite{lim2026scalable}.
GenTact procedurally generates robot-conforming capacitive skins with integrated conductive routing and uses simulated contacts to refine sensor placement~\cite{kohlbrenner2025gentact}, while GenTact-Prox extends this pipeline to proximity sensing~\cite{kohlbrenner2026gentactprox}.
\methodname starts from robot geometry and jointly generates the sensing layout, routing, connectorization, mechanical structure, and associated metadata, enabling robot-specific tactile skins.

\textbf{Tactile simulation and manipulation.}
TACTO~\cite{tacto} and Taxim~\cite{taxim} focus on simulating the imaging process inherent in vision-based tactile sensors. DiffTactile~\cite{difftactile}, TacEx~\cite{tacex}, and Taccel~\cite{taccel} additionally implement sensor surface deformation mechanics for robot manipulation. HydroShear further models shear and stick-slip effects for
sim-to-real learning~\cite{hydroshear}. Tactile feedback has been used in fine manipulation and shear-based control~\cite{3dvitac,simshear}. Different to prior work, \simname implements a simulation framework specifically for \coolname sensors using the compiled sensor geometry and bump layout to generate normal and tangential deformation observations.

\textbf{Acquisition and cross-sensor representations.}
High-rate tactile sensing has used event-driven hardware~\cite{NeuTouch,evetac} and adaptive subsampling of dense arrays~\cite{slepyan2024acts}. \xamplingname applies adaptive sampling across arbitrary compiled layouts using manifest-defined valid regions and neighborhoods. 
Cross-sensor tactile representation learning combines aligned synthetic tactile data with real-world multimodal data to learn sensor-agnostic visuo-tactile representations~\cite{sparsh, anytouch,pmlr-v306-zhang26gh}. \transfername instead maps calibrated contact fields through physical coordinates and local frames, without needing paired cross-sensor data.

\section{The OpenTac Ecosystem}
\label{sec:system}

\begin{figure*}[t]
    \centering
    \includegraphics[width=.95\linewidth]{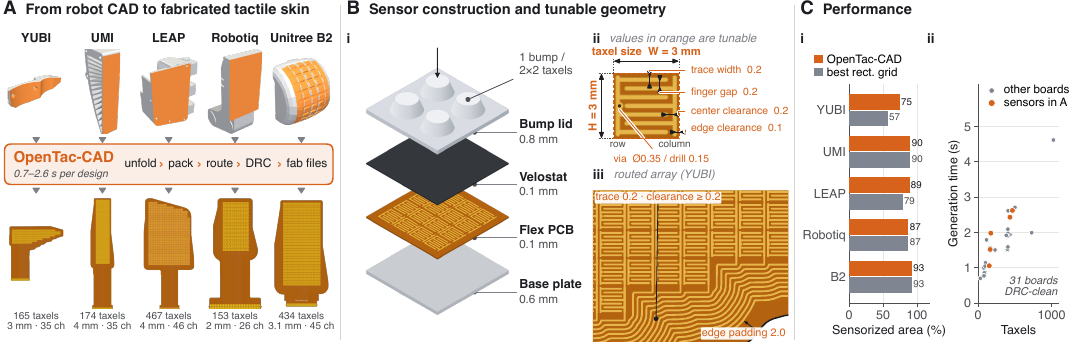}
    \caption{\textbf{\cadname creates manufacturable tactile skins for 2D/3D shapes.}
    (A) Selected surfaces (orange) on YUBI, UMI, LEAP Hand, Robotiq 2F-85, and Unitree B2 models and the generated FPCBs; labels give taxel count, pitch, and row-plus-column readout channels.
    (B) (i) Sensor stack: a printed bump lid places one bump over each $2\!\times\!2$ taxel group above the piezoresistive film and two-layer FPCB. (ii) One generated taxel and (iii) a routed corner of the YUBI board; values in orange are compiler inputs, shown at the YUBI settings, that can be changed to generate a different sensor.
    (C) Sensorized area versus the largest inscribed rectangular array at the same pitch, and generation time versus taxel count for all 31 fabricated boards, all with zero DRC violations.}
    \label{fig:cad}
    \vspace{-0.2cm}
\end{figure*}

OpenTac introduces a full-stack ecosystem for piezoresistive tactile skins in robot control and learning applications. 
The OpenTac tactile sensor comprises a structured deformable overlay, which redistributes tangential loading across a small grid of underlying taxels. This enables the estimation of both normal and shear forces over the piezoresistive film.

All \methodname components use a shared sensor description: the \emph{manifest}. The ecosystem comprises four distinct modules:
\begin{enumerate}
    \item \cadname, which designs a sensor given a 2D outline or a developable 3D shape, producing the fabrication files and the manifest;
    \item \simname, a simulation toolkit built on MuJoCo to simulate sensor readings;
    \item \transfername, a module to enable cross-sensor geometry transfer; and
    \item \xamplingname, an adaptive readout module enabling kilohertz-rate sampling for sparse contact acquisition.
\end{enumerate}
Together, these enable \methodname to be directly deployed across highly versatile traditional, learning-based, and interactive robotics applications.

\subsection{\methodname: Normal and Shear Sensor design}
\label{sec:sensing}

We choose piezoresistive sensing for its low material cost, simple readout circuitry, and compatibility with standard flexible-PCB (fPCB) fabrication. 
The OpenTac sensor stack (Fig.~\ref{fig:cad}B) comprises three primary components: a single fPCB (bottom), a piezoresistive film (middle), and a printed mechanical cover (top).
At each row--column intersection on the fPCB, interdigitated copper fingers form one taxel.
Any applied pressure on the sensor decreases the local resistance on the piezoresistive film and is converted by a readout circuit into an output voltage to estimate normal force.
The top surface of the sensor is a deformable bump strip, based on~\cite{bump-strip-reference}, which is where sensor-object interactions occur.
Each bump sits atop a small window (e.g., $2 \times 2$) of taxels.

For three-axis force sensing, each elastomeric bump covers a local group of taxels, with the number and arrangement of taxels determined by the sensor geometry. For a $2\!\times\!2$ configuration, let the baseline-subtracted signals be $z_{uv}$ for $u,v\in\{-1,+1\}$, with signs denoting each taxel's offset from the bump center.
Three-axis tactile readings are then: %
\begin{equation}
    h_z=\sum_{u,v}z_{uv},\qquad
    h_x=\sum_{u,v}u z_{uv},\qquad
    h_y=\sum_{u,v}v z_{uv}.
    \label{eq:shear}
\end{equation}
Normal loading produces a near-uniform pressure distribution over all taxels underneath a bump, resulting in a common-mode response $h_z$. Tangential loading tilts the bump, resulting in a pressure, and by extension, resistance differential across the underlying taxels, producing differential responses $h_x$ and $h_y$.
Optionally, we may use an affine map to calibrate sensor readings $[h_x,h_y,h_z]$ to force units $[F_x,F_y,F_z]$.
The manifest defines the grouping and signs for each compiled array.
This design choice enables a high degree of customization, by leveraging automated CAD and circuit design principles to parametrize sensor density, electrode layout, and mechanical transduction, while retaining the sensing, assembly, and readout processes.

\subsection{\cadname: Automatic sensor design}
\label{sec:workflow}

\looseness=-1
To deploy \methodname sensors on a new robot, surface, or end-effector, we need to first custom-design a sensor that fits the surface. With existing sensors, customization is  challenging, necessitating expertise across electronic circuit design, electrode spacing, cable routing, mechanical design and modeling, and more. To address this, and to enable high degree of customizability, we build \cadname. Given a 2D or 3D shape, our compiler deterministically generates the board geometry. Users can customize sensing regions, taxel footprint, pitch, electrode and clearance rules, cable, and connector.

\begin{figure*}[t!]
    \centering
    \includegraphics[width=0.9\linewidth]{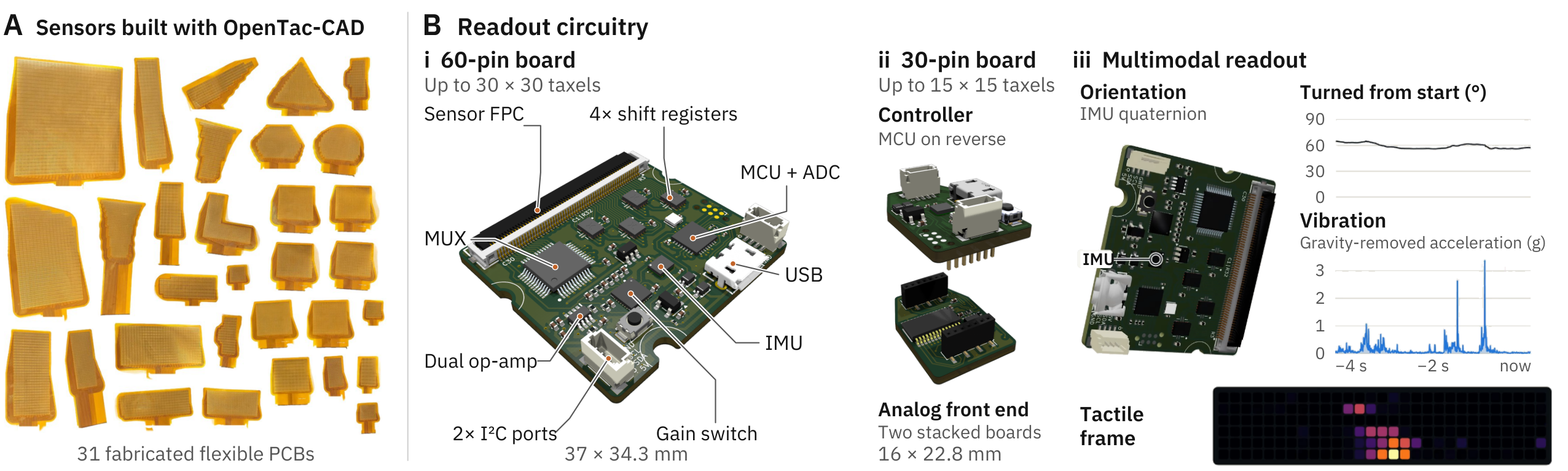}
    \caption{\textbf{A.} 31 sensors in various form factors designed with \cadname and fabricated. All sensors were tested and work. \textbf{B.} We provide two \textbf{readout board} designs, varying in the size and the number of sensor wires they support. Each board contains filtering and readout circuitry, an IMU, and a microcontroller. It can be plugged directly into a USB port to stream data.}
    \label{fig:otc-fabricated-sensors-and-readout}
    \vspace{-0.2cm}
\end{figure*}

\textbf{Sensor outline extraction.}
\cadname supports sensor design for a user-drawn 2D input region, a 2D outline imported from a vector drawing (DXF), or a 3D mesh (STL) where users specify the surface(s) to be sensorized. 
In the case of a 3D mesh, surfaces are first unfolded to 2D triangle by triangle across shared edges. Since each triangle's flat shape follows exactly from its edge lengths, developable surfaces such as cylinders, cones, and faceted parts flatten without distortion, and doubly curved selections that cannot be flattened are detected and reported. We additionally support an agentic outline extraction mode, where a large language model processes an input text prompt and designs a 2D outline. Extracted outlines from this step are passed along to the next step for wiring.

\textbf{Taxel size, pitch, and placement.}
The default size across most \coolname sensors is 4 mm taxels at 4.2 mm centre-to-centre spacing (also referred to as the \textit{pitch}). In our tests, we fabricated sensors ranging from 4 mm to 1 mm taxel sizes. Reducing the pitch increases the number of taxels within a given area and provides finer spatial sampling; increasing taxel size increases the signal measured. 
Sensor width and height can be independently configured if needed.
Once taxel size and pitch are determined, we populate taxels all along the area to be sensorized, accounting for any additional free space required to route the wires originating from the taxels. We also allow users to specify additional design constraints, such as additional boundary space on the sensor edge, or the length and shape of the sensor \textit{tail} (i.e., the side of the sensor where the wires are all routed to, and connected to the readout board).

\textbf{Readout boards.}
We provide two readout board designs: a 60-pin board for sensors with up to 30 rows and 30 columns, and a compact 30-pin pair for up to 15 rows and 15 columns. Both use the same firmware codebase and USB interface, with the sensor-specific layout supplied through its manifest.
One half of the connector drives up to 32 lines through shift registers; the other senses up to 32 lines through a multiplexer and an inverting amplifier that holds the sensed line at virtual ground to suppress crosstalk.
The compiler assigns one electrode set to each connector half and allocates pins outward from the center. 
This enables skins of varying sizes to be plugged into the same readout board.

\textbf{Contour-following routing.}
Rows and columns lie on opposite faces of the fPCB, so routing each face is a planar single-layer problem.
Every net follows a span of the sensor-array contour, split into hops around concavities. Spans on a face must be nested or disjoint, so unavoidable crossings are detected a priori.
Routes are processed from the shortest to the longest span. Each route follows a clearance-offset contour around the array and previously routed copper, avoiding an expensive routing search. Every route is checked for clearance before acceptance.
This estimates the required routing band for each sensor outline.

\textbf{Sensor Manifest}
The manifest describes the sensor's mechanical and electronic construction, and enables interoperability across all tools in the \coolname ecosystem.
For every taxel $i$, it stores the physical center $\mathbf{p}_i$, row--column scan indices $(r_i,c_i)$, and footprint parameters, together with the sensing boundary, manufactured outline, connector pin map, and the recipe that generated the board.
Bump strips additionally store each taxel's mechanical group $b(i)$. 
Separating the geometric information from raster scan ordering is important for irregular skins, because the wires are optimized for routing, and that order could be different from that of the physical coordinates. Fig.~\ref{fig:otc-fabricated-sensors-and-readout} shows 31 sensors that were designed entirely in \cadname and fabricated using a flexPCB printing service.

\begin{figure*}[t!]
    \centering
    \includegraphics[width=\linewidth]{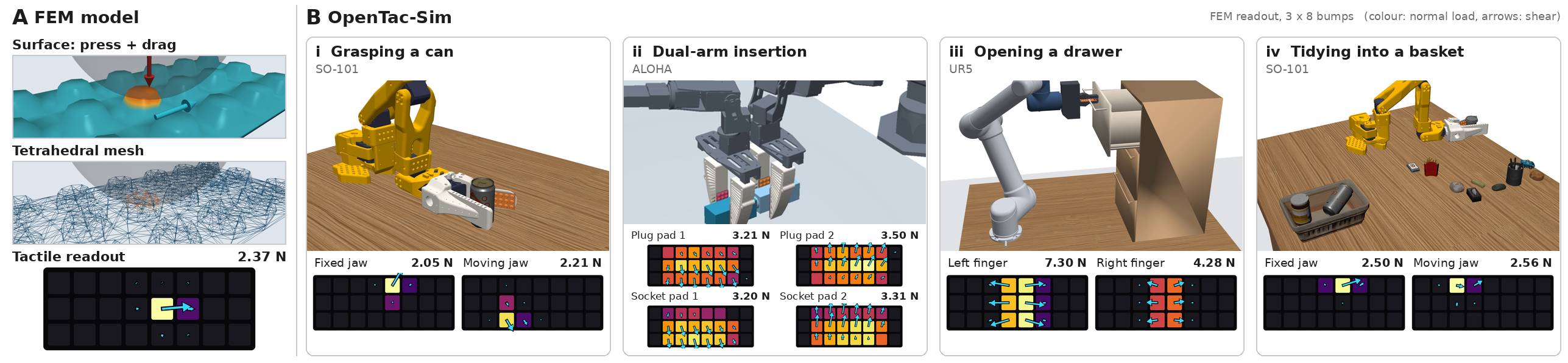}
    \caption{\textbf{\simname} enables the simulation of \coolname sensors, ensuring that normal and shear forces are accurately modeled. We use an FEM model that reads the sensor manifest created by \cadname and supports the simulation of sensors of arbitrary shapes and resolutions. \textbf{A.} FEM model uses a tetrahedral mesh to simulate bump deformation and computes normal and shear forces using Eq.~\ref{eq:shear}. \textbf{B.} \simname includes support for multiple robots from the MuJoCo menagerie, across several contact-rich tasks.}
    \label{fig:opentac-sim-approach-section}
\end{figure*}

\subsection{\simname: Simulating \methodname sensors}
\label{sec:simulation}

\looseness=-1
\simname provides a set of accurate simulation methods for \coolname sensors. The bump-covered sensors that measure both normal and shear forces are simulated using an FEM model, while smooth sensor skins that only measure normal force are simulated using a reduced foam-contact model. Sensor geometry and output formats are directly instantiated from the sensor manifest, enabling \simname to support the simulation of sensors of arbitrary shapes and resolutions.

\textbf{Sensor model}:
The manifest file describes sensor geometry and is used in the initialization of multiple tactile \textit{pads}. Each pad is converted into a tetrahedral mesh and is simulated as a linear-elastic FEM body in MuJoCo, with its base pinned onto the sensor base plate (or a robot gripper, as applicable). Upon physical contact with an object, the tactile pads deform based on the contact geometry and the load (Fig.~\ref{fig:opentac-sim-approach-section}A).
We retain the original surface mesh for collision evaluation.
At each bump, we compute normal and shear forces in the local tactile pad frame and interpolate them to taxel resolution.

\textbf{Robot and task simulation}:
We support the simulation of multiple robots from MuJoCo menagerie. New robots and tasks can be easily instantiated. The current release of the simulator only supports stable simulation of sensor-object interactions with rigid objects.

\textbf{Tactile readout}: To simulate the effects that the sensor stack and circuitry would have on taxel readings, we also model interaction between the tactile pads, the base, and the electrode footprints. Starting from the FEM state, we compute the elastic force exerted by each tactile pad on the base, map it onto the electrode footprints, and apply our normal and shear force estimation method from Eq.~\ref{eq:shear}. We run an automated calibration step per sensor to remove loads originating from the pad's own scaled mass under gravity and arm acceleration.

\textbf{Smooth sensor simulation}:
The simulation also supports smooth sensor skins using a
reduced foam-contact model. We model the foam as a Winkler layer,
$p=k(\delta-z)_+$, solve $\delta$ to match the applied normal force,
smooth the resulting pressure field with a Gaussian of width $\sigma$,
and integrate it over the manifest taxels before applying the target
calibration. 

\subsection{\transfername: Cross-sensor transfer}
\label{sec:transfer}

\transfername maps calibrated normal-contact fields between skins in physical coordinates.
With source and target taxel cells $A_i$ and $B_j$, contact-frame alignment $T$, and source calibration $C_s$,
\begin{equation}
    M_{ji}=\frac{|T(A_i)\cap B_j|}{|A_i|},\qquad
    \hat{F}_j=\sum_i M_{ji}C_s(s_i),
    \label{eq:transfer}
\end{equation}
and target counts are predicted through the inverse target calibration, $\hat{s}_j=C_t^{-1}(\hat{F}_j)$, with its noise floor and saturation applied.
Force is conserved when the target covers the contact.
To compute this mapping, we only need the calibration map (an affine transform) per-sensor, and for the contact alignment to be known. 

\subsection{\xamplingname: Adaptive Acquisition}
\label{sec:xampling}

Piezoresistive sensor arrays are commonly read via \textit{raster scanning}. We read taxels left-to-right and top-to-bottom, by energizing one row and column at a time.
For $N$ taxels, this raster strategy costs $N$ samples even when contact occupies only a small region.
This results in two unintended consquences. First, readout rates slow down with increase in the area or resolution of a tactile skin. Second, fast contacts that typically occur in under 10 milliseconds go undetected by the sensor.
To mitigate these, \xamplingname assigns a budget of $M\ll N$ samples per frame, using the binary-quadtree adaptive subsampling of~\cite{slepyan2024acts}.
We allocate one quarter of this per-frame sampling budget to search over the full mask, and the remainder to follow active regions detected in the previous frame.
This leverages the idea that contacts tend to be sparse, local, and temporally coherent. The approach works on any kind of sensor layout (rectangular, concave, curved-part, etc.).
In the worst-case (i.e., when every single taxel experiences contact), this gracefully degrades to have identical complexity as raster scanning.

\begin{figure}[t!]
    \centering
    \includegraphics[width=\linewidth]{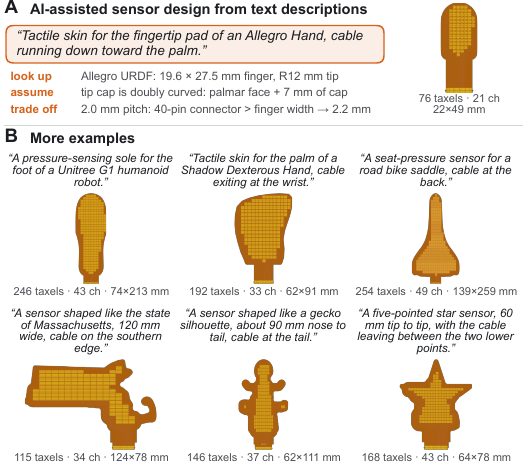}
    \caption{\textbf{Tactile skins from one-sentence language descriptions.}
    A coding agent (Claude Opus~5) generated each board design from the quoted prompt using \cadname and web search.
    (A) Logged decisions for an Allegro Hand fingertip: a dimension looked up online, an assumption about the tip geometry, and a pitch change forced by connector width.
    (B) Further designs; the bottom row used outline-growth mode at the operator's request.
    Labels give taxel count, routed row-plus-column nets, and board size. These examples evaluate automated design generation, not fabrication.}
    \label{fig:agentic}
    
\end{figure}

\section{Experimental Evaluation}
\label{sec:evaluation}

We experimentally evaluate every component of the \methodname ecosystem, including the compiler, sensor characterization, acquisition, transfer, and applicability to contact-rich control and manipulation policy learning.

The key experimental findings are that:
\begin{enumerate}
    \item \cadname auto-generates designs that result in higher taxel coverage, without any design violations
    \item \methodname sensors provide accurate shear and normal force estimates, measured against reference commercial 3-axis force sensors
    \item \simname faithfully simulates the sensors, as quantified by aligned real-and-simulated sensor data
    \item \methodname computes accurate shear for downstream contact-rich control and manipulation policies
    \item \transfername enables transferring normal forces across sensor designs
    \item  \xamplingname reads out at kilohertz-rates, measuring fast contacts that raster scanning methods miss
    \item High Dynamic Range (HDR) readout re-reads saturated taxels at lower gains within each frame, measuring loads beyond the range of a fixed gain

\end{enumerate}

\begin{figure}[t!]
    \centering
    \includegraphics[width=\linewidth]{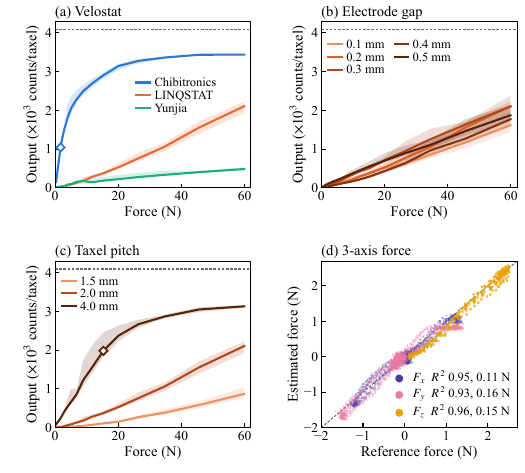}
    \caption{\textbf{Sensor calibration.}
(a--c) Per-taxel response to normal loading against a load cell: median and interquartile range per force bin; diamonds mark saturation onset (5\% of pressed taxels at full scale); dashed lines mark the 12-bit ADC limit.
(a) Three films (2.0~mm pitch, 0.2~mm gap).
(b) Electrode gaps of 0.1--0.5~mm (LINQSTAT, 2.0~mm pitch).
(c) Pitches of 1.5, 2.0, and 4.0~mm (LINQSTAT, 0.2~mm gap).
(d) In-sample three-axis calibration of a bumped $4\!\times\!4$ patch against an OptoForce reference; the dashed line is $y=x$, and legends give $R^2$ and RMSE.}
    \label{fig:calibration}
\end{figure}

\subsection{Evaluating auto-generated robot skin designs}
\label{sec:cad_results}

We compiled and fabricated 31 boards spanning robot-derived and synthetic outlines, taxel counts, pitches, and electrode gaps.
Generation was run three times per design on an Apple M4 Pro laptop; all reported boards completed with zero design-rule violations, and every fabricated board was functional.
Figure~\ref{fig:cad}A highlights five embodiments: a YUBI finger pad, UMI soft-gripper finger, LEAP Hand palm, Robotiq 2F-85 finger pad, and Unitree B2 foot.
They contain 153--467 taxels at 2--4~mm pitch and require only 35--46 electrical channels (an $R\!\times\!C$ matrix needs $R+C$ lines).

To quantify the benefit of shape-aware packing, we compare the active taxel-cell area with the largest rectangular grid of the same pitch that fits inside each target region.
\cadname covers 75\%, 90\%, 89\%, 87\%, and 93\% of the YUBI, UMI, LEAP, Robotiq, and B2 regions, respectively (Fig.~\ref{fig:cad}C).
The rectangular baseline covers 57\%, 90\%, 79\%, 87\%, and 93\%.
As expected, both methods agree on nearly rectangular regions; the compiler gains most on tapered or irregular ones, improving YUBI and LEAP by 18 and 10 percentage points.
Across the five highlighted designs, compilation from outline to fabrication files takes 0.7--2.6~s.

\textbf{Design generation from natural language.}
We pair OpenTac-CAD with a VLM agent (Claude Opus 5) to take in a natural language description specifying arbitrary design requirements and constraints.
The VLM agent then constructs the sensing outline, optionally web-searching as needed, and selects taxel size, pitch, and board mode, the same inputs a user enters in the interface.
This design is then passed to the OpenTac-CAD compiler, which handles all subsequent steps, including electrode and clearance rules, packing, readout-aware wiring, routing, and design-rule checks.
The agent can call the compiler repeatedly, for example to sweep pitch, before selecting a design. Using OpenTac-CAD and web search, all nine agent sessions produced DRC-clean layouts (76--254 taxels, median 6.5~min), each recording dimensional sources and design assumptions. (Fig.~\ref{fig:agentic}).

\begin{figure}[t!]
    \centering
    \includegraphics[width=\linewidth]{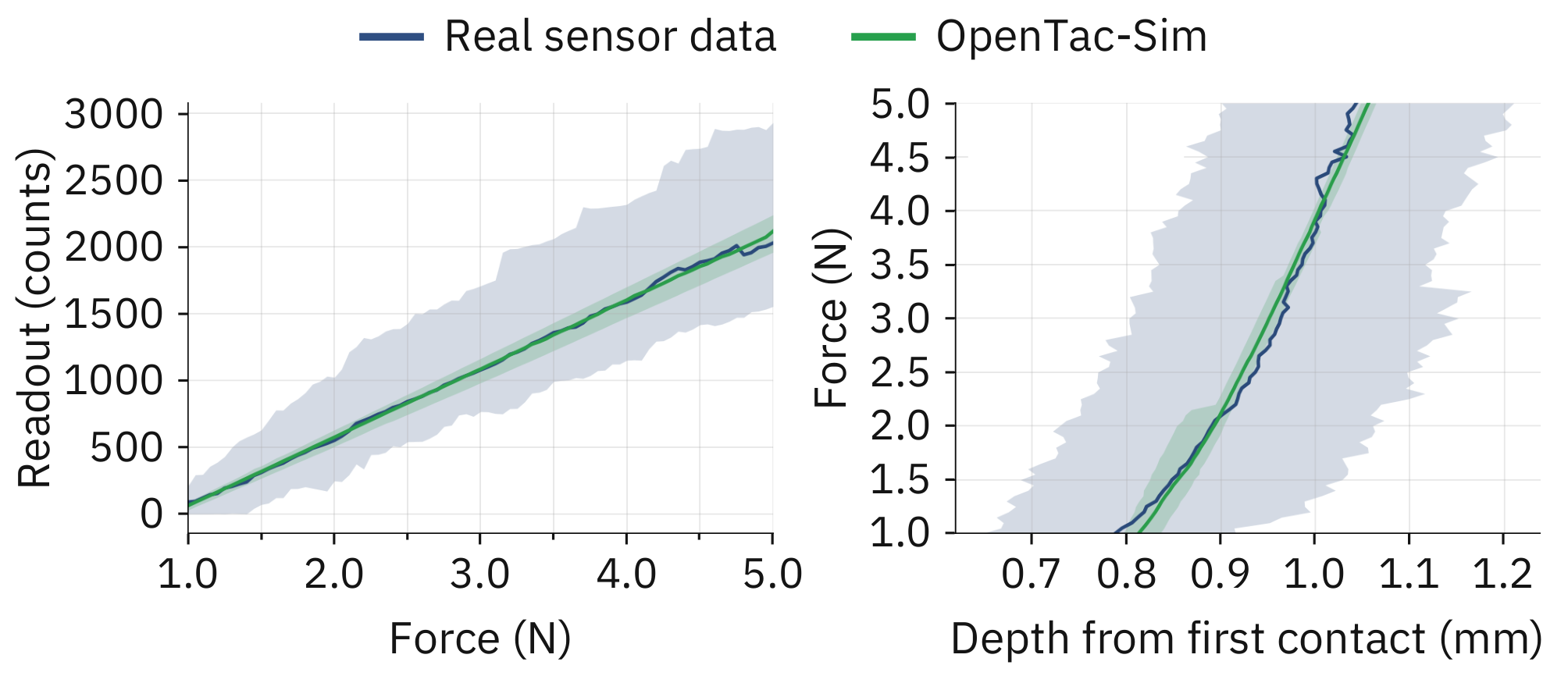}
    \caption{\textbf{\simname} faithfully simulates our shear sensor. We collect data by pressing the bump strip on top of a calibrated sensor with a force-controlled UR5e robot with forces ranging from 1 N to 5 N. We reproduce the forces in simulator and compare alignment between simulation vs real-world force readings and contact depths.}
    \label{fig:opentac-sim-vs-real-plot}
\end{figure}

\subsection{Shear, material, and geometry Characterization}
\label{sec:calibration_results}

We conduct multiple studies to characterize the sensor and answer two questions: 
\begin{enumerate}
    \item How do each of the design choices (piezoresistive material, electrode geometry) impact force response? 
    \item How accurate are the normal and shear force estimates from \coolname compared to reference 3-axis force sensors?
\end{enumerate}
The experiments all involve pressing a square sensor array pressed against a load cell, while varying one design variable at a time. The results are shown in Fig.~\ref{fig:calibration} a--c.

\textbf{Choice of piezoresistive film}:
We compare 3 commercially available piezoresistive films -- Chibitronics, LINQSTAT, and Yunjia. Each film exhibits different sensitivity and saturation behavior. At 2.0~mm pitch and 0.2~mm gap, Chibitronics saturates at low force, LINQSTAT does not reach the saturation criterion through 60~N, and Yunjia produces the lowest ADC response over the tested force range.

\textbf{Electrode geometry}:
Electrode gap has a modest effect on response, whereas larger pitch increases the per-taxel signal. Taxel pitch therefore affects both spatial resolution and sensitivity.
For instance, 2 mm taxel pitch with a Chibitronics piezoresistive has a 60 N force range, but increasing the taxel pitch to 4 mm saturates the sensor earlier (at about 20 N).

\textbf{Force estimation accuracy}:
We conduct this experiment with our \textit{default} configuration, which is a $4\!\times\!4$ bump strip array using a LINQSTAT piezoresistive sheet with a 4.0~mm taxel pitch and 0.2~mm electrode gap. The sensor is first calibrated with a reference force measured by the load cell. We then apply a range of reference forces across the X, Y, and Z axes on the bumps and log sensor measurements and reference forces measured by a load cell. 
We measure correlation between \coolname and the reference force sensor by computing $R^2$. The estimation method in Eq.~\ref{eq:shear} achieves $R^2$ of 0.95, 0.93, and 0.96 and RMSE of 0.11, 0.16, and 0.15~N for $F_x$, $F_y$, and $F_z$ against OptoForce (Fig.~\ref{fig:calibration}d).

\begin{figure}[t!]
    \centering
    \includegraphics[width=\linewidth]{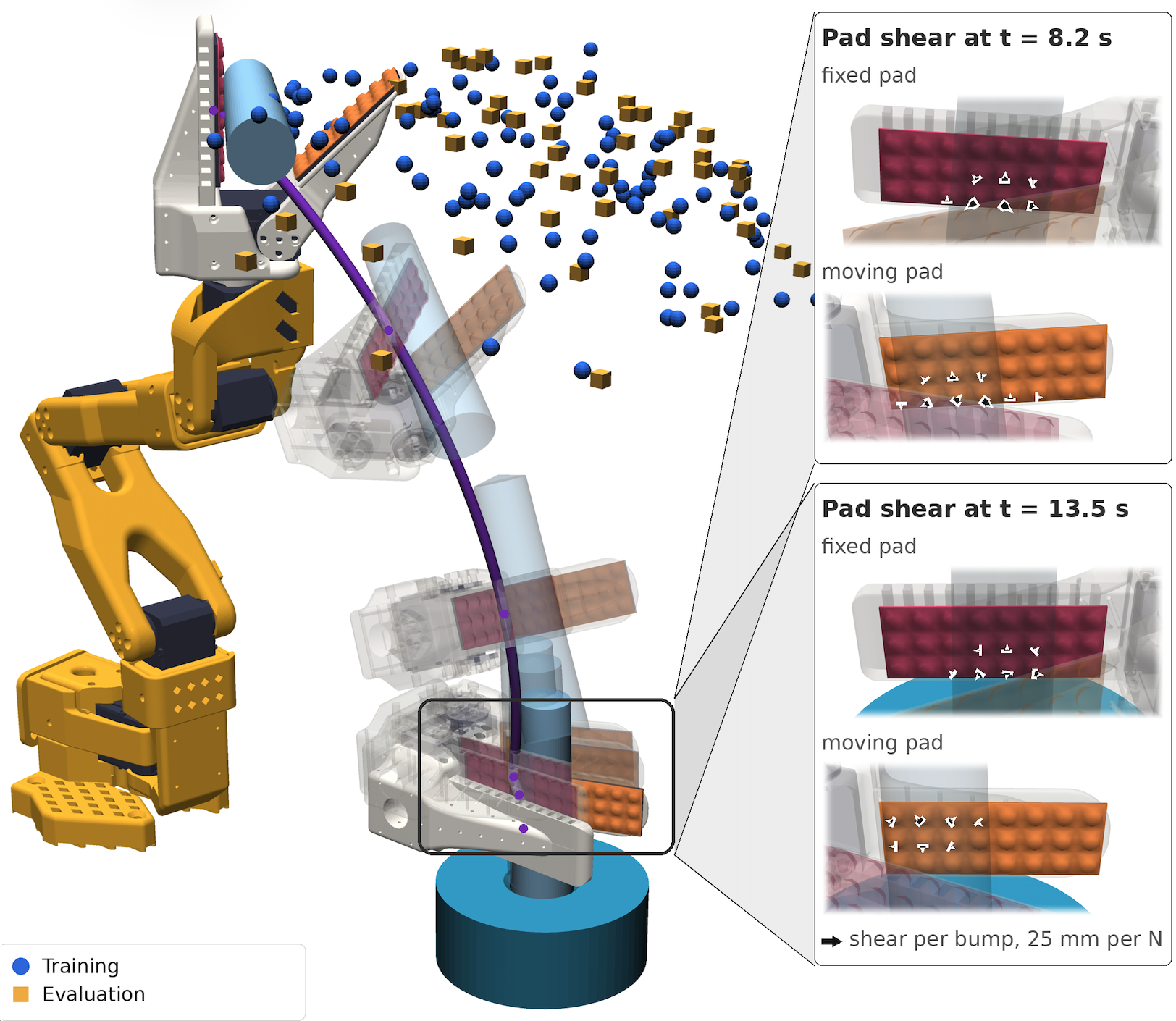}
    \caption{SO-101 peg insertion with deformable tactile pads.
Blue spheres and orange cubes mark training ($n=100$) and held-out
evaluation ($n=50$) start poses, respectively. The purple curve shows
an expert trajectory, with translucent gripper configurations at
multiple timestamps; the dashed gray curve shows an alternative
demonstration.
Inset shows close-ups of the pads during insertion. Arrows show per-bump tangential displacement.}
    \label{fig:opentac-sim-policy-learning}
\end{figure}

\begin{figure*}[t]
    \centering
    \includegraphics[width=\linewidth]{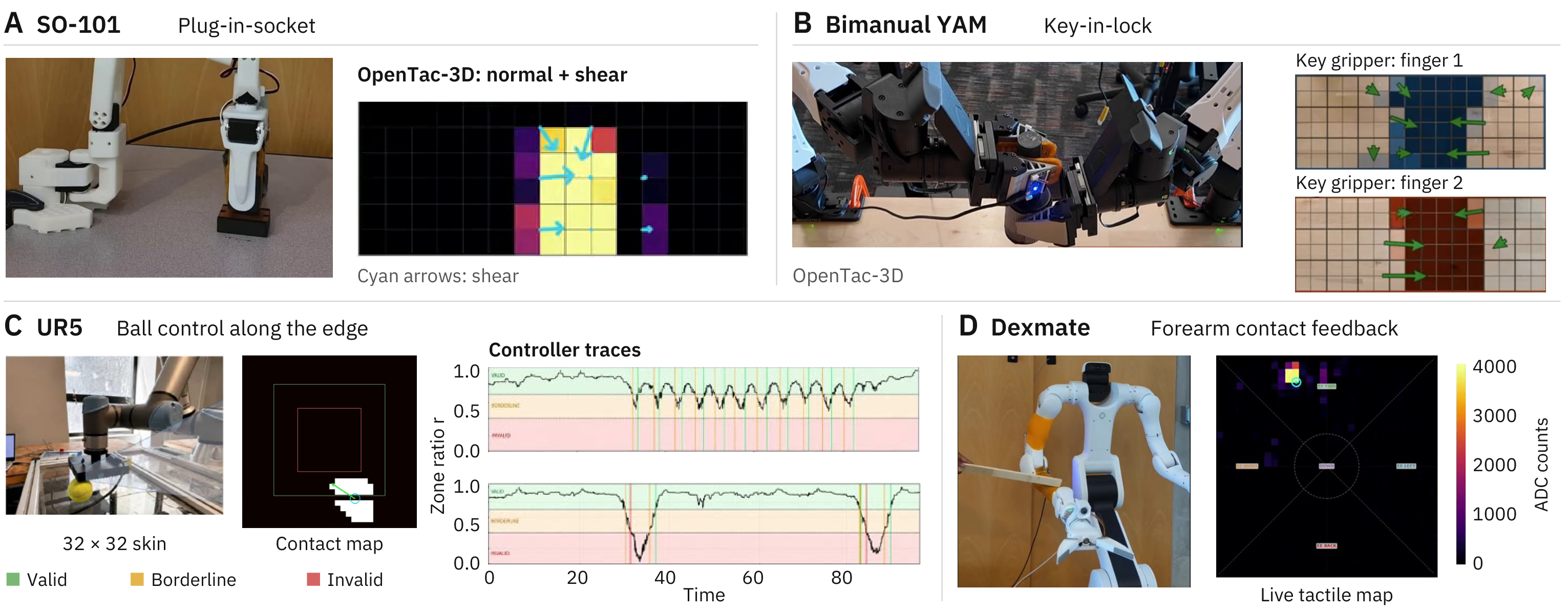}
    \caption{\textbf{Robot deployments.} OpenTac sensors deployed on four physical robot platforms to support policy learning, contact-rich control, and interactive applications. \textbf{A.} \textit{Plug-in-socket insertion} on an SO-101 robot arm where shear sensing is central to task success. \textbf{B.} \textit{Key-in-lock} insertion using a custom-designed cylindrical locking mechanism. \textbf{C.} \textit{Rolling a ball} using a shear-controller on the UR5e robot. \textbf{D.} \textit{Interactive impedance control} on the Dexmate Vega-1 Pro robot. Tactile skin on the robot's forearm senses and moves in the direction of a push.}
    \label{fig:robot_exp}
\end{figure*}

\begin{figure*}
    \centering
    \includegraphics[width=\linewidth]{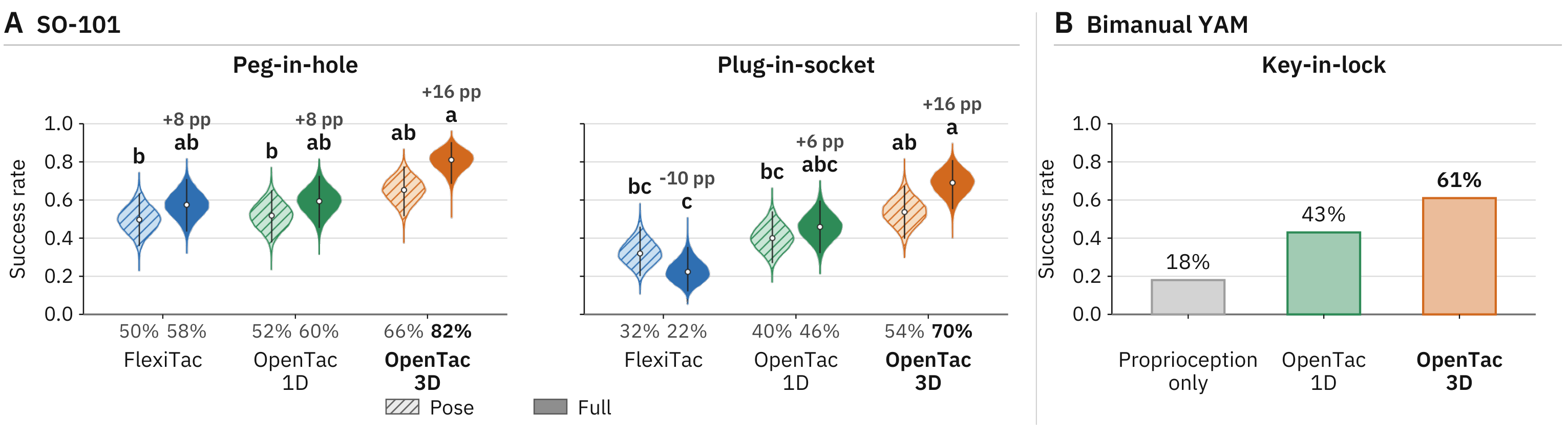}
    \caption{\textbf{Tactile policy learning:} \textbf{A.} \textit{Peg-in-hole} and \textit{Plug-in-socket} insertion on an SO-101 robot arm. Success rates are reported over 50 trials for each condition. ``Pose'' uses joint angles only; ``full'' adds tactile observations; neither uses vision. OpenTac-1D senses normal pressure; OpenTac-3D adds shear. We evaluate FlexiTac~\cite{flexitac} as a baseline normal force sensor. To account for each sensor resulting in slightly different grip friction (e.g., due to bump strips) and demonstration distributions, we train and evaluate a proprioception-only variant for each sensor as a baseline. We use compact letter display (CLD), based on statistical significance testing protocols from~\cite{lbmtri2025}. Two violins that do not share a letter are (statistically) significantly different ($>90\%$ confidence interval). \textbf{B.} \textit{Key-in-lock} insertion task on bimanual YAM robot arms. Success rates are reported across 25 trials. Across all tasks, shear-sensing variants (OpenTac-3D) achieve higher success rates.}
    \label{fig:tactile-policy-learning-quantitative}
\end{figure*}

\subsection{Shear sensor simulation}
\label{sec:results_opentac_sim}

To validate whether the simulation faithfully models a real \coolname shear sensor, we collected sensor measurements by pressing the bumps with a force-controlled UR5e robot end-effector. We vary the forces from 1 N and 5 N in increments of 0.5 N. We then repeat the experiment in simulation, where we press down a bump with exactly the same sequence of forces as we did in the real-world. We calibrate the real-world sensor so we can map the sensor readings to force units. We then compute two correlation plots, shown in Fig.~\ref{fig:opentac-sim-vs-real-plot}. We notice that simulated sensor measurements are extremely well correlated with the mean value of real sensor data. Measurements from real sensor exhibit signal-dependent noise that generally increases with the force value. We also find a strong correlation between the contact depth and force across real-world and \coolname. We tested the viability of the simulator for manipulation policy learning by training a peg insertion policy in the simulator, and show example qualitative results in Fig.~\ref{fig:opentac-sim-policy-learning}.

\subsection{Tactile Policy Learning}
\label{sec:results_tactile_policy}

We evaluate OpenTac for real-world policy learning on two contact-rich manipulation tasks on two robot platforms: an SO-101 robot arm and a bimanual YAM robot. 

\textbf{SO-101 robot arm}:
On the SO-101 robot arm, we evaluated the \emph{peg-in-hole} from LeFlexiTac~\cite{leflexitac} and \emph{plug insertion} from TacOBench~\cite{tacobench}.
In the first, the robot inserts a peg already held in its gripper into a slot.
Plug insertion is a harder task and requires sub-centimeter alignment of the electrical contacts with the socket. 
We compare FlexiTac~\cite{flexitac}, OpenTac-1D (normal-force only; no bump strip), and OpenTac-3D (normal and shear forces; with the bump strip). 
The two OpenTac conditions use the same skin and readout board and differ only in the cover. 
For each sensor, we train an action chunking transformer (ACT) that takes tactile signals as input in addition to joint angles (referred to as ``full'') and a variant that uses joint angles only (referred to as ``pose''). See 
Fig.~\ref{fig:tactile-policy-learning-quantitative}. 
Each policy is trained on  100 demonstrations and evaluated over 50 trials. 
We also train a per-sensor proprioception-only baseline to account for any variation in sensor hardware and in demonstrator trajectory distribution. In particular, adding bumps changes contact mechanics as well as sensitivity to shear.

\begin{figure*}[t]
    \centering
    \includegraphics[width=\textwidth]{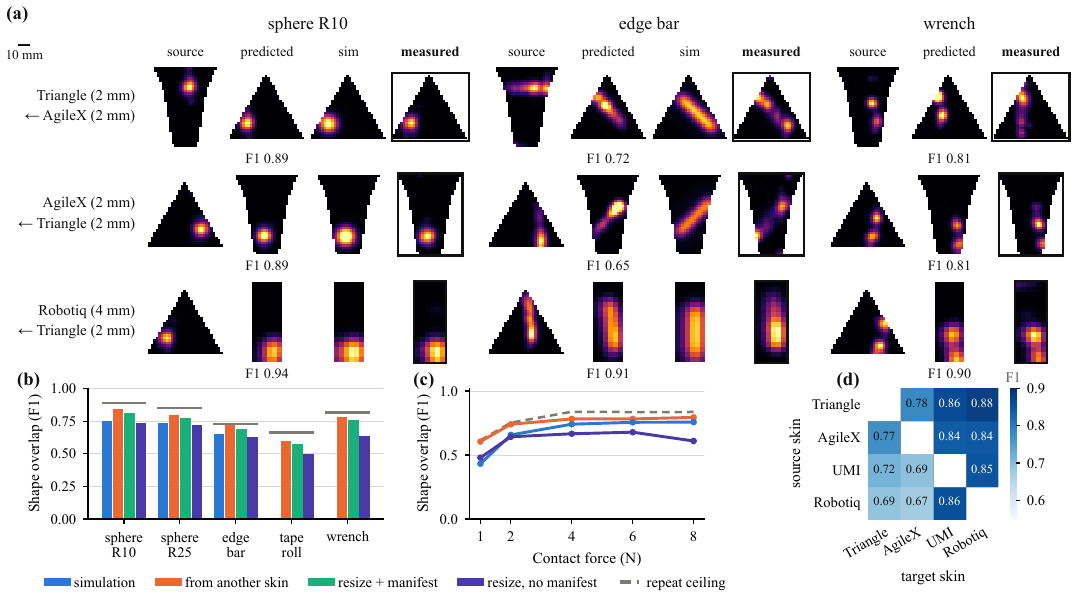}
    
    \caption{\textbf{Predicting tactile contact across skins and from a reduced foam model.}
    (a) Median-performing presses at $\geq4$~N; rows are target $\leftarrow$ source at a shared physical scale (bar: 10~mm). Predicted, simulated, and measured panels share the target's peak scale; sources use their own.
    (b) Median shape F1 by object; grey marks same-sensor repeatability.
    (c) F1 versus calibrated force.
    (d) Transfer F1 for each ordered skin pair on co-visible support. All predictors receive contact pose. Simulation covers printed indenters only; Robotiq simulation exemplars are in-sample and excluded from aggregates.}
    \label{fig:transfer}
    
\end{figure*}

OpenTac-3D-full achieves the highest observed success rate on both tasks. On peg-in-hole, tactile observations improve success from 66\% to 82\%, compared with 50\% to 58\% for FlexiTac and 52\% to 60\% for OpenTac-1D. 
In plug insertion, OpenTac-3D-full succeeds in 70\% of trials against 54\% for its pose-only baseline(+16 percentage points).
Normal force may not distinguish alignment with the socket from contact against its surrounding surface. 
Neither normal-only tactile policy improves on its corresponding proprioception-only baseline in these trials. 
The higher success rate with OpenTac-3D suggests that shear is essential for the plug insertion task, as also independently reported in~\cite{tacobench}. Additionally, as shown in Fig.~\ref{fig:robot_exp} (top-left), the normal force maps are nearly identical regardless of whether the plug is in the air or inserted in the socket. It is only the shear estimates that vary significantly and provide cues for task success.

\textbf{Bimanual YAM robots}:
We evaluate a \textit{key-in-lock} insertion task on bimanual YAM robot arms. We design a cylindrical locking mechanism where a \textit{key} with an extruded \textit{slot} needs to be twisted and inserted into a \textit{lock}. The key will only fit into the lock at a specific configuration. Like the previous experiments, no vision is assumed available. For this task, one of the arms holds the lock and the other attempts to twist or rotate and insert the key. Success rates over 25 trials are reported in Fig.~\ref{fig:tactile-policy-learning-quantitative}. This is an extremely challenging task for behavioral cloning methods that solely rely on proprioception. It is evident from the low success rates (18\%) for the `proprioception-only' variant. Shear sensing is pivotal here, enabling the detection of whether the key is in the slot, on the edge, or not in contact with the lock at all. The `OpenTac 3D' variant that uses shear improves success rate to 62\%, when compared to the `OpenTac 1D` variant that uses only normal force and achieves a success rate of 43\%.

\begin{figure*}[t]
    \centering
    \includegraphics[width=\textwidth]{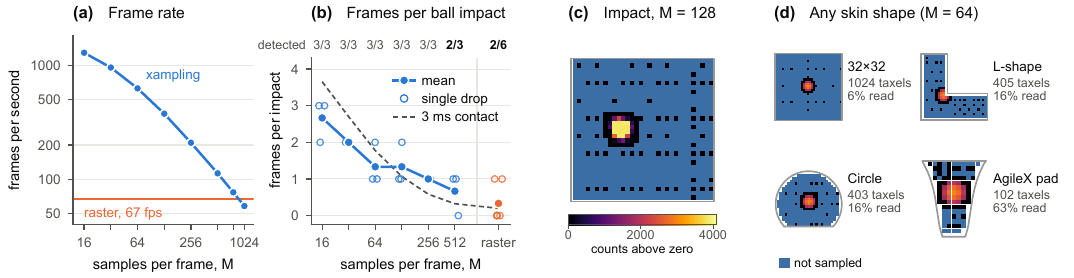}
\caption{\textbf{Adaptive partial scanning.}
(a) Frame rate versus per-frame sampling budget $M$ on a $32\!\times\!32$ skin; full raster scanning reads all 1024 taxels at 67~frames/s.
(b) Frames capturing a tennis-ball impact from a 30~cm drop (50\% full-scale threshold). Open circles are trials and filled circles means; the dashed curve gives the frames spanned by a 3~ms contact at each frame rate; detection counts appear above.
(c) Representative impact at $M=128$.
(d) Firmware sampling patterns at $M=64$ on four skin shapes (not to scale). Color is ADC counts above baseline; black taxels were sampled without contact, and blue taxels were not sampled.}
    \label{fig:xampling}
    
\end{figure*}

\subsection{Contact-rich control tasks}

\label{sec:results_control_tasks}

We evaluate OpenTac on two contact-rich control tasks, comparing shear feedback against normal force alone.

\textbf{Ball manipulation.}
A UR5 with a $32\times32$ OpenTac sensor on its end-effector rolls a ball along the end-effector's outer edges without ball-position feedback (Fig.~\ref{fig:robot_exp}, UR5 panel), regulating toward a target shear vector along the edge.
The shear controller recovers from the ``Borderline'' to the ``Valid'' region more reliably than normal force alone.

\textbf{Bimanual grasp stability under impulsive loading.}
A Dexmate Vega-1 Pro holds a paper cup between two OpenTac probes
under proprioceptive, wrist force/torque, normal-pressure, or aggregate-shear feedback, with identical trajectories, gains, and a 2.5~N nominal grip; the shear controller raises this target before slip.
Table~\ref{tab:grasp_stability} reports trials that keep the cup for 40~s (\emph{Held}) and net cup displacement (\emph{Slip}).

\begin{table}[t]
\centering
\caption{Hold stability on a Dexmate Vega-1 Pro. Twenty trials per condition,
Slip is reported as mean\,$\pm$\,std.; Held counts trials reaching 40\,s.
---~dropped, so slip is undefined.}
\label{tab:grasp_stability}
    
\small
\setlength{\tabcolsep}{4pt}
\renewcommand{\arraystretch}{1.15}
\begin{tabular*}{\linewidth}{@{\extracolsep{\fill}}lcc@{}}
\toprule
Feedback & Held $\uparrow$ & Slip (cm) $\downarrow$ \\
\midrule
\multicolumn{3}{@{}l}{\textbf{No load}} \\
Proprioception only & 3 & --- \\
F/T sensor & 20 & $0.15\pm0.08$ \\
OpenTac-1D & 20 & $0.2\pm0.1$ \\
OpenTac-3D & 20 & $\mathbf{0.11\pm0.06}$ \\
\midrule
\multicolumn{3}{@{}l}{\textbf{Load}} \\
Proprioception only & 0 & --- \\
F/T sensor & 9 & $5.5\pm1.3$ \\
OpenTac-1D & 9 & $5.8\pm2.1$ \\
OpenTac-3D & \textbf{16} & $\mathbf{2.6\pm1.2}$ \\
\bottomrule
\end{tabular*}

\end{table}

\subsection{Cross-Sensor Transfer}
\label{sec:transfer_results}

We evaluate four LINQSTAT skins without bumps---Triangle (409 taxels) and AgileX (502) at 2~mm pitch, and UMI (174) and Robotiq (96) at 4~mm---pressing five objects at 95 locations and scoring contacts at 1--8~N using each skin's calibration.
Every predictor receives the contact pose.
We compare calibrated transfer with raw-count area averaging over manifest geometry (``resize + manifest''), bilinear array resizing (``resize, no manifest''), and the foam model of Sec.~\ref{sec:simulation}, scoring shape F1 between the predicted and measured active taxels after a one-pitch blur and a threshold at 25\% of each field's peak.
Transfer reaches a median F1 of 0.750 over 5,290 predictions, versus 0.714 and 0.634 for the two resizing baselines; on support covered by both skins it reaches 0.788, close to the 0.800 same-skin repeatability.
Coarser 4-mm targets score higher than 2-mm targets (0.84--0.88 vs.\ 0.67--0.78; Fig.~\ref{fig:transfer}d), and the foam model reaches 0.727 on the three skins excluded from its fit.
These results cover pose-conditioned normal-contact shape, not force magnitude or sim-to-real policy transfer.

\begin{figure}
    \centering
    \includegraphics[width=\linewidth]{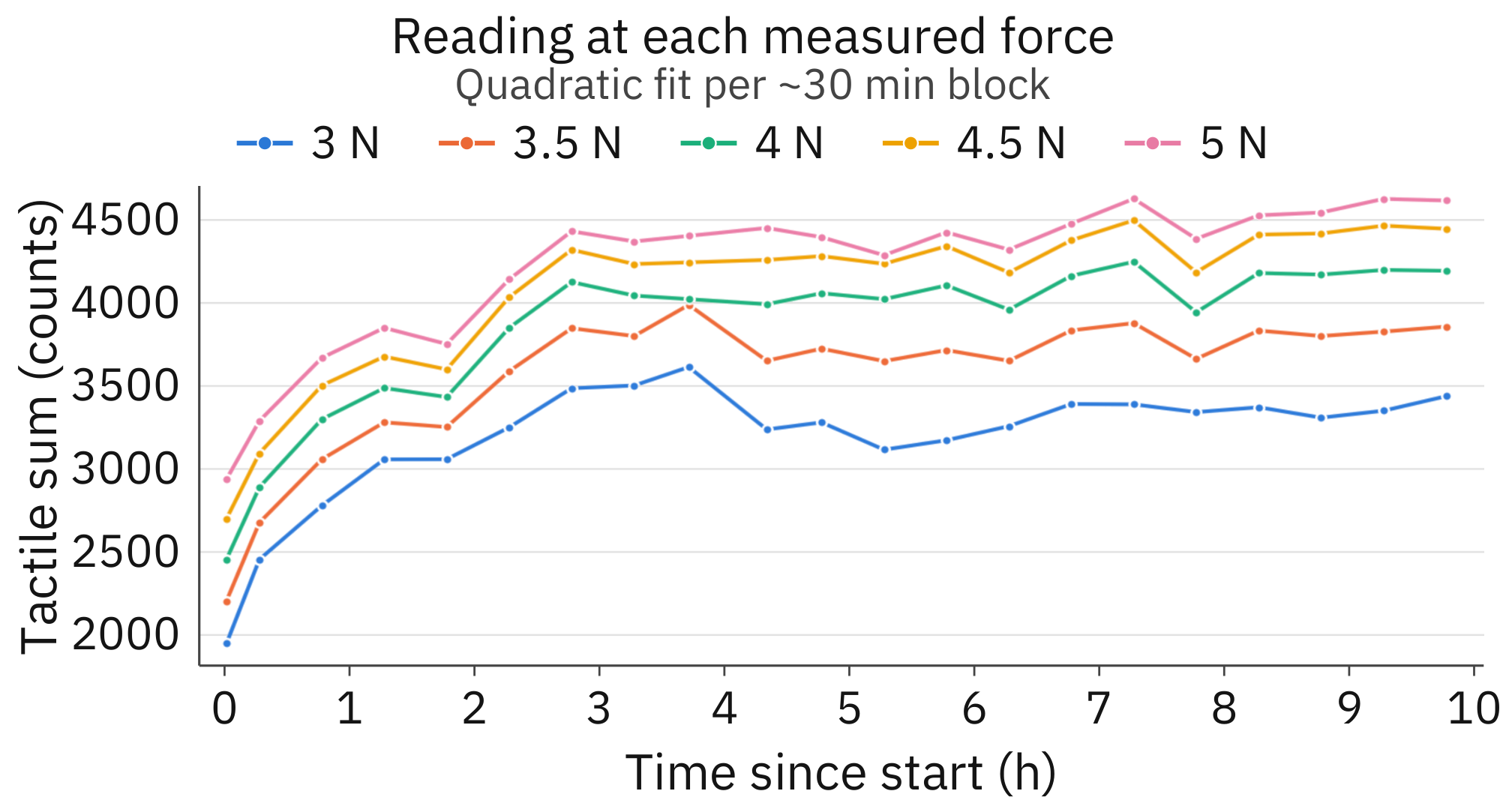}
    \caption{\textbf{Sensor endurance testing}: We continuously indent a bump on a sensor with a robot end effector for 10 hours. Every 30 minutes, we log sensor readings for forces in the range 3 N - 5 N. We find that after the first 2 hours of interaction, the readings stabilize, and maintain a clear relative separation throughout the 10-hour load-cycle test.}
    \label{fig:endurance-test}
\end{figure}

\subsection{High-Speed Adaptive Sampling}
\label{sec:xampling_results}

On a $32\!\times\!32$ skin, raster scanning reads all 1024 taxels at 67 frames/s, while shrinking the adaptive budget raises the rate monotonically to 1286 frames/s at $M=16$, a $19.2\times$ increase (Fig.~\ref{fig:xampling}a).
To evaluate whether partial scanning resolves short-duration contacts, we drop a tennis ball from 30~cm and count frames above 50\% of ADC full scale: raster scanning detects only two of six impacts, each in a single frame, whereas budgets $M=16$--256 detect all three tested impacts per condition and the fastest settings record several frames of the roughly 3~ms event (Fig.~\ref{fig:xampling}b); at $M=512$, two of three impacts are detected as partial scanning approaches the raster limit.
The same firmware runs unchanged on square, L-shaped, circular, and gripper-pad manifests (Fig.~\ref{fig:xampling}d), reading 6\% of the 1024-taxel square and 63\% of the 102-taxel pad per frame at $M=64$, without a hand-coded raster or neighborhood map.

\begin{figure*}[t]
    \centering
    \includegraphics[width=\textwidth]{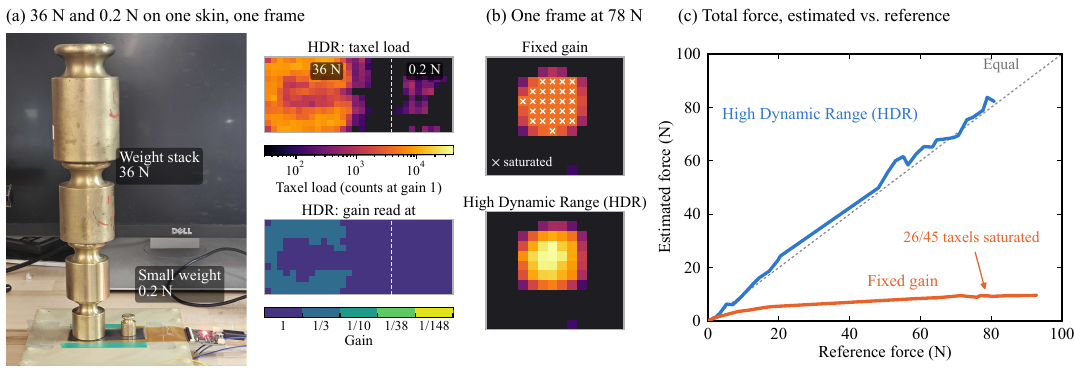}
    \caption{\textbf{High Dynamic Range (HDR) readout} on a 360-taxel skin resting on a reference scale.
    (a)~A 3.7~kg weight stack (36~N, left) and a 20~g weight (0.2~N, right; nominal masses) on opposite ends of the skin: photo, and one HDR frame with the gain each taxel was read at. The dashed line marks the gap between the two contacts. HDR reads the stack at gains of 1 and 1/3 and the small weight at gain~1; at a fixed gain, 99 of the stack's 195 taxels in contact saturate.
    (b)~A plunger pressing the center of the skin, one frame at 78~N in each mode. At a fixed gain, 26 of the 45 taxels in contact saturate ($\times$); HDR re-reads only those taxels, at gains of 1/3 and 1/10, within the same frame. Color in (a) and (b) is the taxel load in counts at gain~1.
    (c)~Force estimated by the skin versus the scale for the plunger in (b): median per 2~N of reference force over all loading and holding samples. Both modes use one calibration, fitted to the same scale in a separate session.}
    \label{fig:hdr}
    \vspace{-0.2 cm}
\end{figure*}

\subsection{High Dynamic Range (HDR) Readout}
\label{sec:hdr_results}

Most existing tactile sensors are configured to read either heavy or light touches, but not both simultaneously on the same contiguous sensor. 
This setting, commonly referred to as High-Dynamic Range (HDR), is natively enabled by \methodname using a per-taxel, customizable amplifier gain setting.
Using the switchable feedback resistance of the readout board's sensing amplifier, each frame is initially read at the highest gain setting.
Taxels that reach a clip threshold just below saturation are re-read, at successively lower gains (1/3, 1/10, 1/38, and 1/148). Each taxel is reported at the highest gain that does not saturate it.
This extends the measurable (i.e., dynamic) range by up to $148\times$ while retaining full sensitivity to light contact.

With a 3.7~kg weight stack (36~N) on one end and a 20~g weight (0.2~N) on the other (Fig.~\ref{fig:hdr}a), HDR mode reads the heavier weight stack at gains of 1 and 1/3 and the small weight at full gain. 
Were these to be read with a fixed gain, 99 of the stack's 195 taxels in contact saturate (i.e., a third of its signal would be lost).
With a plunger pressing the center of the skin against a reference scale (Fig.~\ref{fig:hdr}b,c), a fixed gain saturates 26 of the 45 taxels in contact at 78~N, and the estimated force levels off near 9~N.
HDR re-reads the contact at gains of 1/3 and 1/10 and follows the scale up to the largest tested load, 80~N.
Above 20~N, the median estimated-to-reference ratio is 1.04, against 0.13 at a fixed gain.
The measured gain steps agree with the nominal feedback-resistance ratios to within 1\%.
Importantly, HDR mode does not add any additional sensing delays when it is not needed (i.e., when no taxels saturate).
Each lower-gain pass adds a fixed 0.3--0.5~ms to switch the gain, and each re-read adds 15~$\mu$s, as much as reading one taxel in a full scan (re-reading 25 taxels over two passes adds 1.2~ms).

\subsection{Endurance (duty cycle) testing}
\label{sec:endurance_testing}

We also carried out endurance testing on \coolname sensors. Using a UR5e robot with an indenter on its end-effector, we constantly press on top of a single bump for 10 hours at a stretch. At every 30-minute interval, we exert forces ranging from 3 N - 5 N (in increments of 0.5 N), and plot the sensor readouts in Fig.~\ref{fig:endurance-test}.
We notice a gradual increase in the force response after the first 2 hours of use, upon which the sensor readings stabilize.
However, throughout the 10-hour test window, we observe that the force readings are distinctly separable in relative magnitudes.
Therefore, either a \textit{warmup} phase for the sensor at the start of a usage session, or using relative sensor measurements as opposed to absolute ADC readings is recommended for consistent results.

\FloatBarrier
\section{Conclusion}
\label{sec:discussion}

We have presented \methodname, an open ecosystem spanning robot surface geometry, fabricated tactile skins, manifest-driven simulation, cross-sensor transfer, high-speed and high dynamic range acquisition.
To the best of our knowledge, \methodname is the first such sensor to support all these features in a single customizable sensor.
Several limitations remain. The flexible PCBs are not stretchable, and mesh-derived layouts require surfaces that can be flattened into developable patches. 
Also, piezoresistive sensing introduces hysteresis, drift, and temperature dependence. 
Adaptive acquisition trades spatial completeness for update rates, while HDR readout trades update rates for measurable range.
We primarily focus on normal and shear force estimation (the sensors support twist estimation, at the cost of spatial resolution).
We release all hardware, firmware, software, and data that help in reproducing the sensor.

\section*{Acknowledgement}
The authors acknowledge in-kind support from Dexmate Inc. and Toyota Research Institute for the robot equipment used in the project. 
KMJ, HB, AB, and GH were supported by a DexMate research grant program.
AS thanks Analog Devices, Inc. for financial support.
Addison Johnson designed the lock-key fixtures used in the bimanual policy learning experiments.
Reyhaneh Behravesh and Chuhan Zhang participated in early brainstorming meetings.
Junjun Chen helped cure silicone for bump designs that were used in early prototypes.
Jonathan Yao contributed helpful initial discussions on a broad range of topics involving mechanical fabrication. 
AS also thanks his postdoctoral advisor, Pulkit Agrawal, for helpful discussions. 

\begingroup
\setlength{\bibsep}{2pt}
\bibliographystyle{unsrtnat}
\bibliography{references}
\endgroup
\clearpage
\newpage
\section*{Contribution Statements}

\textbf{Ariel Slepyan} conceived and initiated the OpenTac project and led its overall development. AS developed the core tactile sensing architecture, designed and fabricated the sensor hardware and readout electronics, and developed the acquisition firmware. AS developed OpenTac-CAD, OpenTac-Transfer, and OpenTac-Xampling, and co-developed OpenTac-Sim. AS established the initial working system and coordinated the broader collaborative effort. AS conceived and designed experiments, prepared figures, and led the writing and preparation of the manuscript.

\textbf{Geeta Chandra Raju Bethala} was a core contributor to OpenTac-CAD, alongside AS. GCRB also ran robot experiments on the DexMate Vega Pro and produced several qualitative results for this work.

\textbf{Jiafeng Gu} led the development of OpenTac-Sim and produced multiple quantitative and qualitative results to assess the efficacy and accuracy of OpenTac-Sim.

\textbf{Anagha Rajendra Dangle} coordinated sensor design, placement, accuracy, and ensured experiment design consistency across multiple experiments for this work, ranging from policy learning to characterization to contact-rich control.

\textbf{Yinzhe Zhou} and \textbf{Tuhin Basu} led the sensor characterization and calibration experiments. YZ and TB also conducted load-cycle testing experiments. TB additionally led the contact-rich control experiments.

\textbf{Hongyuan Fu} and \textbf{Xinyi Yin} led the SO-101 policy learning experiments. XY instrumented sensors, designed/fabricated test fixtures, and
HF collected data and trained and evaluated policies. 
\textbf{Md Faizal Karim} led the bimanual manipulation experiments using the YAM robots. \textbf{Jade Sim} designed and fabricated a tactile glove using OpenTac sensors.

\textbf{Homanga Bharadhwaj} helped shape and advise the policy learning experiments, participated in the weekly discussions and planning for the project's experiments, and participated in paper writing and designing visuals.  

\textbf{Anand Bhattad} led weekly project discussions and shaped the vision for the OpenTac-Sim. AB was also heavily involved in sensor characterization and policy learning experiments, and wrote/edited significant portions of this manuscript.

\textbf{Krishna Murthy Jatavallabhula} was involved in \textit{nearly every} experiment run for the paper. KMJ advised on experiment design, overall framing/scope, and wrote/edited significant portions of this manuscript and release materials.

\textbf{Gregory Hager} and \textbf{Nitish V. Thakor} provided research supervision, scientific guidance, and strategic advice throughout the project. They shaped the project's research direction, experimental scope, and overall presentation.

\end{document}